\documentclass{article}
\usepackage{spconf,amsmath,graphicx}
\usepackage{amssymb}
\usepackage{booktabs}
\usepackage{multirow}
\usepackage[english]{babel}
\usepackage{amsthm}
\usepackage{setspace}

\def\Vec#1{\textsf{\boldmath $#1$}}
\def\ceq{\mathop{=}^c}
\def\E{{\mathbb E}}
\def\C{{\mathbb C}}
\def\T{{\textsf{T}}}
\def\H{{\textsf{H}}}
\def\Normal{{\mathcal{N}}}
\def\Diag{\mathop{\mathrm{diag}}\nolimits}

\newcommand{\refeq}[1]{(\ref{eq:#1})}

\newcommand{\refsec}[1]{Section \ref{sec:#1}}

\newcommand{\reffig}[1]{Fig. \ref{fig:#1}}

\newcommand{\reftab}[1]{Table \ref{tab:#1}}

\newcommand{\citess}[2]{\cite{#1}--\cite{#2}}

\title{Fast MVAE: Joint separation and classification of mixed sources based on multichannel variational autoencoder with auxiliary classifier}
\name{Li Li$^1$, Hirokazu Kameoka$^2$, Shoji Makino$^1$ 
\thanks{This work was supported by JSPS KAKENHI Grant Number 17H01763 and 18J20059, and SECOM Science and Technology Foundation.}}
\address{$^1$ University of Tsukuba, Japan \\
$^2$ NTT Communication Science Laboratories, NTT Corporation, Japan }
\begin{document}
%
\maketitle
\begin{abstract}
This paper proposes an alternative algorithm for multichannel variational autoencoder (MVAE), a recently proposed multichannel source separation approach.
While MVAE is notable in its impressive source separation performance, the convergence-guaranteed optimization algorithm and that it allows us to estimate source-class labels simultaneously with source separation, there are still two major drawbacks, i.e., the high computational complexity and unsatisfactory source classification accuracy. 
To overcome these drawbacks, the proposed method employs an auxiliary classifier VAE, an information-theoretic extension of the conditional VAE, for learning the generative model of the source spectrograms. 
Furthermore, with the trained auxiliary classifier, we introduce a novel algorithm for the optimization that is able to not only reduce the computational time but also improve the source classification performance. 
We call the proposed method “fast MVAE (fMVAE)”.
Experimental evaluations revealed that fMVAE achieved comparative source separation performance to MVAE and about 80\% source classification accuracy rate while it reduced about 93\% computational time.

\end{abstract}
\begin{keywords}
Multichannel source separation, multichannel variational autoencoder, auxiliary classifier, source classification
\end{keywords}
\section{Introduction}
\label{sec:intro} 
Blind source separation (BSS) is a technique for separating out individual source signals from microphone array inputs when both the sources and the mixing methodology are unknown.
The frequency-domain BSS approach allows us to perform instantaneous mixture separation and provides the flexibility of utilizing various models for the time-frequency representations of source signals.
For example, independent vector analysis (IVA) \cite{kim2006independent,hiroe2006solution} solves frequency-wise source separation and permutation alignment simultaneously by assuming that the magnitudes of the frequency components originating from the same source tend to vary coherently over time.
Multichannel extensions of non-negative matrix factorization (NMF), e.g., multichannel NMF (MNMF)  \cite{ozerov2010multichannel,sawada2013multichannel} and independent low-rank matrix analysis (ILRMA)  \cite{kameoka2010statistical,kitamura2016determined}, provide an alternative solution to jointly solving these two problems by adopting NMF concept for the source spectrogram modeling. 
Specifically, the power spectrograms of the underlying source signals are approximated as the linear sum of a limited number of basis spectra scaled by time-varying amplitudes. 
It is noteworthy that IVA is equivalent to ILRMA in a particular case where only a single basis spectrum consisting of ones is used for each source signal. 
In this point of view, ILRMA can be interpreted as a generalized method of IVA that incorporates a source model with stronger representation power, which has shown to significantly improve the source separation performance \cite{kitamura2016determined}. 

Motived by this fact and the high capability of deep neural networks (DNNs) to spectrogram modeling, some attempts have recently been made to use DNNs for source models instead of the NMF model 
\citess{nugraha2016multichannel}{seki2018}.
Multichannel variational autoencoder (MVAE) \cite{kameoka2018semi} is one of these methods that achieved great success in multi-speaker separation tasks.
MVAE trains a conditional VAE (CVAE) \cite{kingma2014semi,sohn2015learning} using power spectrograms of clean speech samples and the corresponding speaker ID as auxiliary label inputs so that the trained decoder distribution can be used as a universal generative model of source signals, where the latent space variables and the class labels are the unknown parameters. 
At separation phase, MVAE iteratively updates the separation matrix using iterative projection (IP) method \cite{ono2011stable} and the unknown parameters of source generative model using backpropagation. 
The separated signals are obtained by applying the estimated separation matrix to the observed mixture signals. 
This optimization algorithm is notable in that the convergence to a stationary point is guaranteed and it allows estimating the source-class labels simultaneously with source separation. 
However, there are two major limitations.
Firstly, the backpropagation needed for each iteration causes the optimization algorithm highly time-consuming, which can be troublesome in practical applications. 
Secondly, the encoder and decoder in a regular CVAE are free to ignore the class labels by finding networks that can reconstruct any data without using the additional information.
In such a situation, the additional class labels will have limited effect on the spectrogram generation, 
which therefore leads to an unsatisfactory source classification result as we will show it in \refsec{experiment}.

To address these limitations, this paper proposes ``fast MVAE (fMVAE)'' that employs an auxiliary classifier VAE (ACVAE) \cite{kameoka2018acvae} for learning the generative distribution of source spectrograms and adopts the trained auxiliary classifier to the optimization at separation phase.

\section{MVAE for determined multichannel source separation}
\label{sec:MVAE}
\subsection{Problem formulation}
Let us consider a determined situation where $I$ source signals are captured by $I$ microphones. 
Let $x_i(f, n)$ and $s_j(f, n)$ denote the short-time Fourier transform (STFT) coefficients of the signal observed at the $i$-th microphone and the $j$-th source signal, 
where $f$ and $n$ are the frequency and time indices respectively.
We denote the vectors containing $x_1(f, n), \ldots, x_I(f, n)$ and $s_1(f, n), \ldots, s_I(f, n)$ by 
\begin{align}   
	\Vec{x}(f, n) &= [x_1(f, n), \ldots, x_I(f, n)]^{\T} \in \C^I,  \\
	\Vec{s}(f, n) &= [s_1(f, n), \ldots, s_I(f, n)]^{\T} \in \C^I,
\end{align}
where $(\cdot)^{\T}$ denotes transpose. 
Under determined situation, the relationship between observed signals and source signals can be described as
\begin{align}
	\Vec{s}(f, n) &= \Vec{W}^{\H}(f)\Vec{x}(f, n), 
	\label{eq:demixing} \\
	\Vec{W}(f) &= [\Vec{w}_1(f), \ldots, \Vec{w}_I(f)] \in \C^{I \times I},
\end{align}
where $\Vec{W}^{\H}$ is called the separation matrix. 
$(\cdot)^{\H}$ denotes Hermitian transpose. 

Let us assume that source signals follow the local Gaussian model (LGM), 
i.e., $s_j(f, n)$ independently follows a zero-mean complex Gaussian distribution with variance $v_j(f, n)=\E[|s_j(f, n)|^2]$
\begin{align}
	s_j(f, n) \sim \Normal_{\C} (s_j(f, n)|0,v_j(f,n)).
\label{eq:LGM}
\end{align}
When $s_j(f, n)$ and $s_{j'}(f, n)$$(j \neq j')$ are independent, $\Vec{s}(f, n)$ follows
\begin{align}
	\Vec{s}(f, n) \sim \Normal_{\C}(\Vec{s}(f,n) | \Vec{0}, \Vec{V}(f, n)),
\end{align}
where $\Vec{V}(f, n)=\Diag[v_1(f, n), \ldots, v_I(f, n)]$. 
From \refeq{demixing} and \refeq{LGM}, we can show that $\Vec{x}(f, n)$ follows
\begin{align}
	\Vec{x}(f, n) \sim \Normal_{\C} (\Vec{x}(f, n) | \Vec{0}, (\Vec{W}^{\H}(f))^{-1}\Vec{V}(f, n)\Vec{W}(f)^{-1}).
\end{align}
Hence, the log-likelihood of the separation matrices $\mathcal{W}=\{\Vec{W}(f)\}_f$ and source model parameters $\mathcal{V}=\{v_j(f,n)\}_{j,f,n}$ given the observed mixture signals $\mathcal{X}=\{\Vec{x}(f, n)\}_{f, n}$ is given by
\begin{align}
\label{eq:ll}
	\log p(\mathcal{X}|&\mathcal{W}, \mathcal{V}) \ceq 
	2N \sum_f \log |\det \Vec{W}^{\H}(f)|  \nonumber \\
	&-\sum_{f, n, j} \Big (\log v_j(f, n) + \frac{|\Vec{w}_j^{\H}(f) \Vec{x}(f, n)|^2}{v_j(f, n)} \Big),
\end{align}
where $\ceq$ denotes equality up to constant terms. 
\refeq{ll} will be split into frequency-wise source separation problems if there is no additional constraint imposed on $v_j(f,n)$.   
This indicates that there is a permutation ambiguity in the separated components for each frequency. 

\subsection{Multichannel VAE}
To eliminate the permutation ambiguity during the estimation of $\mathcal{W}$, MVAE trains a conditional VAE (CVAE) to model the complex spectrograms $\Vec{S}=\{s(f, n)\}_{f, n}$ of source signals so that the spectral structures can be captured. 
CVAE consists of an encoder network $q_\phi (\Vec{z}|\Vec{S}, c)$ and a decoder network $p_\theta(\Vec{S}|\Vec{z}, c)$, where the network parameters $\phi$ and $\theta$ are trained jointly using a set of labeled training samples $\{\Vec{S}_m, c_m\}_{m=1}^M$. 
Here, $c$ denotes the corresponding class label represented as a one-hot vector indicating to which class the spectrogram $\Vec{S}$ belongs.
For example, if we consider speaker identities as the class category, each element of $c$ will be associated with a different speaker.
CVAE is trained by maximizing the following variational lower bound 
\begin{align}
\hspace{-20 pt}
	\mathcal{J} (\phi, \theta) = \E_{(\Vec{S}, c)\sim p_D(\Vec{S}, c)}
	&[\E_{\Vec{z}\sim q_\phi(\Vec{z} |\Vec{S}, c)} [\log p_\theta(\Vec{S}|\Vec{z}, c)]
	 \nonumber \\
	&~~~-KL[q_\phi(\Vec{z}|\Vec{S}, c) || p(\Vec{z})]],
\end{align}
where $\E_{(\Vec{S}, c)\sim p_D(\Vec{S}, c)} [\cdot]$ denotes the sample mean over the training examples $\{\Vec{S}_m, c_m\}_{m=1}^M$ and $KL[\cdot||\cdot]$ denotes Kullback–Leibler divergence.
Here, the encoder distribution $q_\phi(\Vec{z}|\Vec{S}, c)$ and the prior distribution of the latent space variable $p(\Vec{z})$ are expressed as Gaussian distributions
\begin{align}
	q_\phi(\Vec{z}|\Vec{S}, c) &= \Normal (\Vec{z}|\Vec{\mu}_\phi (\Vec{S}, c), \Diag (\Vec{\sigma}^2_\phi(\Vec{S}, c))),  \\
	p(\Vec{z}) &= \Normal(\Vec{z}|\Vec{0}, \Vec{I}),
\end{align}
where $\Vec{\mu}_\phi(\Vec{S}, c)$, $\Vec{\sigma}^2_\phi(\Vec{S}, c)$ are the encoder network outputs. 
The decoder distribution $p_\theta(\Vec{S}|\Vec{z}, c)$ is defined as a zero-mean complex Gaussian distribution and a scale parameter $g$ is incorporated to eliminate the energy difference between the normalized training data and test data. 
Hence, the decoder distribution is expressed as  
\begin{align}    
	p_\theta(\Vec{S}|\Vec{z}, c, g) &= 
	\prod_{f, n} \Normal_{\C}(s(f, n)|0, v(f, n)), 
	\label{eq:decoder}\\
	v(f, n) &= g \cdot \sigma_\theta^2(f, n; \Vec{z}, c),
\end{align}
where $\sigma^2_\theta(f,n; \Vec{z}, c)$ denotes the $(f, n)$-th element of the decoder output.
It is noteworthy to mention that the decoder distribution \refeq{decoder} is given in the same form as the LGM \refeq{LGM} so that the trained decoder distribution can be used as a universal generative model with ability to generate complex spectrograms belonging to all the source classes involved in the training examples. 
If we use $p_\theta(\Vec{S}_j|\Vec{z}_j, c_j, g_j)$ to express the generative model of the complex spectrogram of the source $j$, a convergence-guaranteed optimization algorithm can be applied to search for a stationary point of the log-likelihood by iteratively updating (i) the separation matrices $\mathcal{W}$ using iterative projection (IP) method \cite{ono2011stable}, 
(ii) the CVAE source model parameters $\Psi=\{\Vec{z}_j, c_j\}_j$ using backpropagation, and (iii) the global scale parameter $\mathcal{G}=\{g_j\}_j$ with the following update rule
\begin{align}
	g_j \leftarrow \frac{1}{FN} \sum_{f,n} \frac{|y_j(f,n)|^2}{\sigma_\theta^2(f,n;\Vec{z}_j, c_j)},
	\label{eq:scale}
\end{align} 
where $y_j(f,n)=\Vec{w}_j^\H (f)\Vec{x}(f,n)$. 
Note that since the class labels are the model parameters estimated during the optimization,  MVAE is able to perform source classification as well.
%

\section{Proposed method: Fast MVAE}
\label{sec:fMVAE}
While MVAE is notable in that it works reasonably well for source separation and has capability to perform source classification simultaneously, there is still a huge room for improvement in the source classification performance. 
With a regular CVAE imposing no restrictions on the manner how the encoder and decoder may use the class labels, the encoder and decoder are free to ignore $c$ by finding distribution satisfying $q_\phi (\Vec{z}|\Vec{S},c) = q_\phi(\Vec{z}|\Vec{S})$ and $p_\theta(\Vec{S}|\Vec{z}, c) =p_\theta(\Vec{S}|\Vec{z})$. 
As a result, $c$ will have little effect on generating source spectrograms that leads to a limited source classification performance. 
To avoid such situations, this paper proposes using an auxiliary classifier VAE \cite{kameoka2018acvae} for learning the generative distribution $p_\theta(\Vec{S}|\Vec{z}, c)$. 


\subsection{Auxiliary classifier VAE}
Auxiliary classifier VAE (ACVAE) \cite{kameoka2018acvae} is a variant of CVAE that incorporates an information-theoretic regularization \cite{chen2016infogan} to assisting the decoder outputs to be correlated as far as possible with the class labels $c$ by maximizing the mutual information between $c$ and 
$\Vec{S}\sim p_\theta (\Vec{S}|\Vec{z}, c)$ conditioned on $\Vec{z}$.
The mutual information is expressed as
\begin{align}
	&I (c, \Vec{S}|\Vec{z})  \nonumber \\
	&= \E_{c\sim p(c), \Vec{S}\sim p_\theta(\Vec{S}|\Vec{z}, c), c'\sim p(c|\Vec{S})}
	[\log p(c'|\Vec{S})] + H(c),
\end{align}
where $H(c)$ represents the entropy of $c$ that can be considered as a constant term.
However, it is difficult to optimize $I(c,\Vec{S}|\Vec{z})$ directly since it requires access to the posterior $p(c|\Vec{S})$. 
Fortunately, we can obtain a variational lower bound of the first term of $I(c, \Vec{S}|\Vec{z})$ by using a variational distribution $r(c|\Vec{S})$ to approximate $p(c|\Vec{S})$:
\begin{align}
	&\E_{c\sim p(c), \Vec{S}\sim p_\theta(\Vec{S}|\Vec{z}, c), c'\sim p(\Vec{S}|c)} 
	[\log p(c'|\Vec{S})] \nonumber \\
	=& \E_{c\sim p(c), \Vec{S}\sim p_\theta(\Vec{S}|\Vec{z}, c), c'\sim p(\Vec{S}|c)}
	[\log \frac{r(c'|\Vec{S})p(c'|\Vec{S})}{r(c'|\Vec{S})}] \nonumber \\
	\geq& \E_{c\sim p(c), \Vec{S}\sim p_\theta(\Vec{S}|\Vec{z}, c), c'\sim p(\Vec{S}|c)}
	[\log r(c'|\Vec{S})] \nonumber \\
	=& \E_{c\sim p(c), \Vec{S}\sim p_\theta(\Vec{S}|\Vec{z}, c)}
	[\log r(c|\Vec{S})],
	\label{eq:vb}
\end{align}
the equality of which holds if $r(c|\Vec{S})=p(c|\Vec{S})$. 
We therefore can indirectly maximizing $I (c, \Vec{S}|\Vec{z})$ by increasing the lower bound with respect to $p_\theta(\Vec{S}|\Vec{z}, c)$ and $r(c|\Vec{S})$.  
One way to do this involves expressing the variational distribution as a neural network $r_\psi(c|\Vec{S})$ and training it along with $q_\phi(\Vec{z}|\Vec{S}, c)$ and $p_\theta(\Vec{S}|\Vec{z}, c)$. 
$r_\psi(c|\Vec{S})$ is called the auxiliary classifier.  
Therefore, the regularization term that we would like to maximize with respect to $\phi$, $\theta$, $\psi$ becomes
\begin{align}
	&\mathcal{L}(\phi, \theta, \psi) \\
	&= \E_{(\Vec{S}, c)\sim p_D(\Vec{S}, c), q_\phi(\Vec{z}|\Vec{S},c)}
	[\E_{c\sim p(c), \Vec{S}\sim p_\theta(\Vec{S}|\Vec{z},c)} 
	[\log r_\psi(c|\Vec{S})]]. \nonumber
\end{align}
Since the labeled training samples can also be used to train the auxiliary classifier $r_\psi(c|\Vec{S})$, ACVAE also includes the cross-entropy
\begin{align}
	\mathcal{I}(\psi)=\E_{(\Vec{S}, c)\sim p_D(\Vec{S}, c)}[\log r_\psi(c|\Vec{S})] 
\end{align} 
in the training criterion. 
The entire training criterion is thus given by 
\begin{align}
	\mathcal{J}(\phi, \theta) + \lambda_\mathcal{L}\mathcal{L}(\phi,\theta,\psi)
	+\lambda_\mathcal{I}\mathcal{I}(\psi),
	\label{eq:acvae}
\end{align}
where $\lambda_\mathcal{L}\geq 0$ and $\lambda_\mathcal{I} \geq 0$ are weight parameters.  

\subsection{Fast algorithm}
Note that the auxiliary classifier $r_\psi(c|\Vec{S})$ not only assists the encoder and decoder to learn a more disentangled representation, but also provides an alternative to the backpropagation process in the original MVAE optimization, which is able to significantly reduce the computational time. 
We summarize the proposed fast algorithm as follows: 
\begin{enumerate}
\setlength{\itemsep}{0.01pt}
\item Train $\phi$, $\theta$ and $\psi$ using \refeq{acvae}.
\item Initialize $\mathcal{W}$.
\item Iterate the following steps for each $j$:
\begin{enumerate}
\item Update $c_j$ using $r_\psi(c_j|\Vec{S}_j)$.
\item Update $\Vec{z}_j$ using $q_\phi(\Vec{z}_j|\Vec{S}_j, c_j)$.
\item Update $g_j$ using \refeq{scale}.
\item Update $\Vec{w}_j(0), \ldots, \Vec{w}_j(F)$ using IP.
\end{enumerate}
\end{enumerate}

\section{Experiments}
\label{sec:experiment}
To evaluate the effect of incorporating an auxiliary classifier into both the source model training and the optimization process, we conducted experiments to compare the multi-speaker separation performances, source classification accuracies and computational times of fMVAE and the conventional methods, i.e., ILRMA \cite{kameoka2010statistical,kitamura2016determined} and MVAE \cite{kameoka2018semi}. 

\subsection{Experimental conditions}
\label{subsec:conditions}
\begin{figure}[t]
\centering
\includegraphics[width=0.55\linewidth]{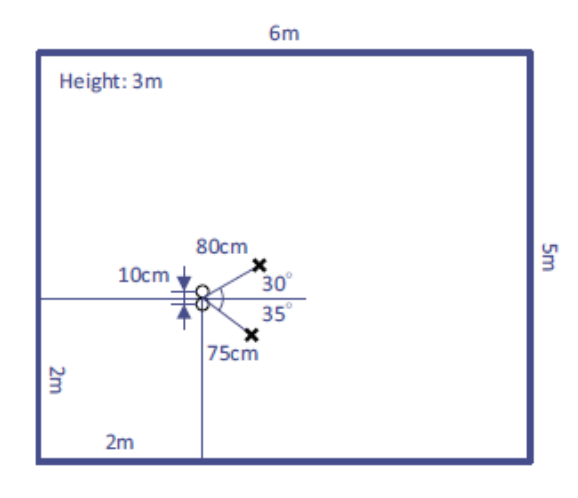} 
\caption{Configuration of the room where $\circ$ and $\times$ represent the position of microphones and sources respectively.}
\label{fig:config}
\end{figure}

\begin{figure*}[t]
\centering
\includegraphics[width=0.8\textwidth]{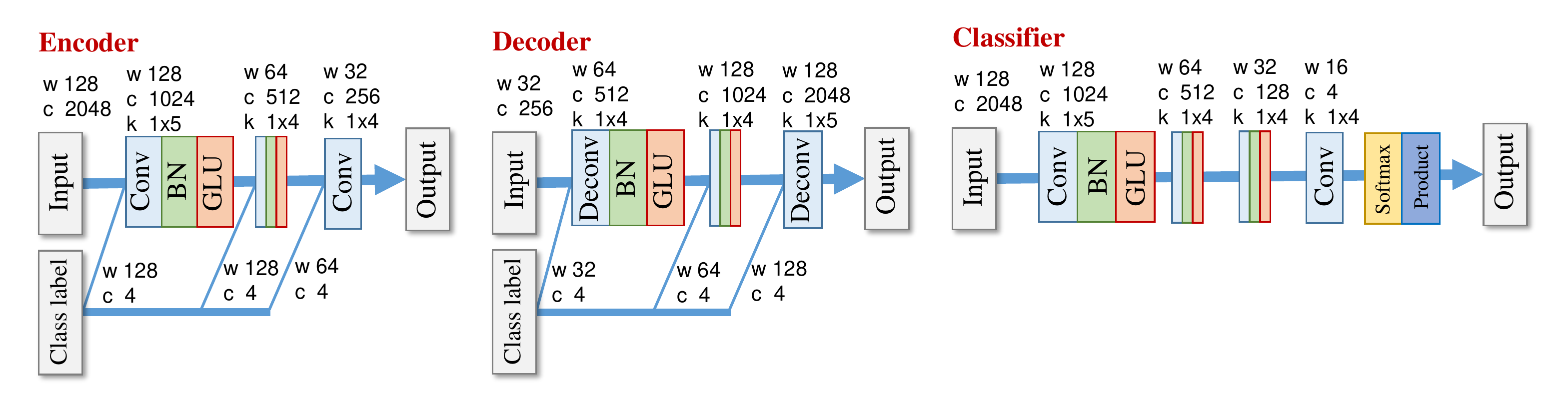} 
\caption{Network architectures of the encoder and decoder used for MVAE and fMVAE and the classifier used for fMVAE. The inputs and outputs are 1-dimensional data, where the frequency dimension of spectrograms is regarded as the channel dimension. ``w'', ``c'' and ``k'' denote the width, channel number and kernel size, respectively. ``Conv'', ``Deconv'', ``BN'' and ``GLU'' denote 1-dimensional convolution and deconvolution, batch normalization, gated linear unit, respectively.}
\label{fig:nn}
\end{figure*}

We excerpted speech utterances from two male speakers (`SM1', `SM2') and two female speakers (`SF1', `SF2') from the Voice Conversion Challenge (VCC) 2018 dataset \cite{lorenzo2018voice}. 
The audio files for each speaker were about 7 minutes long and manually segmented into 116 short sentences, where 81 and 35 sentences (about 5 and 2 minutes respectively) were used as training and test sets, respectively. 
The mixture signals were created by simulated two-channel recordings of two sources where the room impulse responses were synthesized using the image method. 
We tested two different reverberant conditions where the reverberation time ($RT_{60}$) was set at 78 ms and 351 ms, respectively.
\reffig{config} shows the configuration of the room. 
We generated test data involving 4 speaker pairs and 10 sentences for each pair, namely there were totally 40 test signals, each of which was about 4 to 7 seconds long.
All the speech signals were resampled at 16 kHz. The STFT was computed using Hamming window with 256 ms long and window shift was128 ms.

ILRMA was run for 100 iterations and both the proposed method and MVAE were run 40 iterations. 
To initialize $\mathcal{W}$ for MVAE and fMVAE, we used ILRMA run for 30 iterations. 
Adam \cite{kingma2015adam} was used for training CVAE and ACVAE, and estimating the latent variables in MVAE. 
The network architectures used for CVAE and ACVAE is shown in \reffig{nn}. Note that we used the same network architectures of the encoder and decoder for CVAE and ACVAE. 
All the networks are designed as fully convolutional with gated linear units \cite{dauphin2017language} so that the inputs are allowed to have arbitrary lengths.
The programs were run using Intel (R) Core i7-6800K CPU@3.40 GHz and GeForce GTX 1080Ti GPU.
  
\subsection{Results}
\label{subsec:results}
\begin{table}[t]
\centering
\caption{Average SDR, SIR and SAR scores of ILRMA, MVAE and fMVAE. The bold font shows the highest scores.}
\label{tab:scores}
\begin{tabular}{c|ccc}
\hline
\multicolumn{1}{c|}{\multirow{2}{*}{method}} & \multicolumn{3}{c}{$RT_{60}$ = 78 ms}                                		 \\
\multicolumn{1}{c|}{}                  & \multicolumn{1}{c}{SDR {[}dB{]}} & \multicolumn{1}{c}{SIR {[}dB{]}} & \multicolumn{1}{c}{SAR {[}dB{]}} \\
\hline
ILRMA	& 14.8997			& 21.3277		& 18.0584				\\
MVAE 	& 21.5912			& 27.2663 		& \bf{25.1616} 			\\
fMVAE 	& \bf{22.5976} 		& \bf{29.8476} 	& 24.8967 				\\
\hline
\hline
\multicolumn{1}{c|}{\multirow{2}{*}{method}} & \multicolumn{3}{c}{$RT_{60}$ = 351 ms}                                      \\
\multicolumn{1}{c|}{}                  & \multicolumn{1}{c}{SDR {[}dB{]}} & \multicolumn{1}{c}{SIR {[}dB{]}} & \multicolumn{1}{c}{SAR {[}dB{]}} \\
\hline
ILRMA	& 4.6840			& 11.6284 		& 7.2364 			\\
MVAE 	& \bf{8.3157} 	& \bf{18.0834} 	& \bf{9.2206} 	\\
fMVAE 	& 6.7814 			& 15.7728 		& 7.7883 			\\
\hline
\end{tabular}
\end{table}

\begin{table}[t]
\centering
\caption{Computational times of MVAE, fMVAE and ILRMA. MVAE and fMVAE were initialized with run ILRMA algorithm for 30 iterations in CPU and run 40 iterations of the optimization algorithms in CPU or GPU. ILRMA runs 100 iterations in CPU.}
\label{tab:runtime}
\begin{tabular}{lcc}
\hline
		&rumtime/iteration[sec] 	& total [sec] \\
\hline
MVAE  (GPU)	&  6.071632				& 260.5953 		\\
fMVAE (CPU)	&  0.389762				& 21.54129		\\
fMVAE (GPU)	& \bf{ 0.097272}			& \bf{17.56694}	\\
\hline
ILRMA (CPU)	& 0.113571				& 18.38221 		\\
\hline
\end{tabular}
\end{table}

\begin{table}[t]
\centering
\caption{Accuracy rates of source classification obtained with MVAE and fMVAE.}
\label{tab:classification}
\begin{tabular}{ccc}
\hline
&all iterations & final estimation \\
\hline
MVAE & 27.91\% &  37.50\% \\
fMVAE & \bf{78.63\%} & \bf{80.00\%} \\
\hline
\end{tabular}
\end{table}

We calculated the average of the signal-to-distortion ratios (SDR), signal-to-interference ratios (SIR) and signal-to-artifact ratios (SAR) \cite{vincent2006performance} over the 40 test signals to evaluate the source separation performance and measured the computational times.
\reftab{scores} and \reftab{runtime} show the source separation results obtained under different $RT_{60}$ conditions and the computational times of ILRMA, MVAE and fMVAE, respectively. 
fMVAE was about 15 times faster than the original MVAE and even a little faster than ILRMA in GPU machines. 
Furthermore, it is noteworthy that fMVAE achieved comparative source separation performance with the original MVAE.  
As the results show, MVAE and fMVAE significantly outperformed ILRMA in terms of all the criteria of source separation performance, which confirmed the effect of the incorporation of the CVAE source model. 
fMVAE obtained higher SDR and SIR than MVAE under a low reverberant environment, but the performance slightly decreased when $RT_{60}$ became long. 
It is interesting to further compare fMVAE with MVAE in high reverberant environments, which is one direction of our future work.


To evaluation the performance of source classification, we computed the classification accuracy rates over the results estimated in each iteration and only in the final estimation.  
\reftab{classification} shows the results that fMVAE significantly improved source classification accuracy with achieving 80\% accuracy rate. 
Our future work also includes further improving the source classification accuracy.

\section{Conclusions}
\label{sec:conclusions}
This paper proposed fMVAE that (i) uses an auxiliary classifier VAE instead a regular CVAE for learning the generative distribution of source signals; (ii) employs the trained auxiliary classifier and encoder for the optimization. 
fMVAE allows us to significantly reduce the computational time and improve the source classification performance. 
The results revealed that fMVAE achieved about 15 times faster than the original MVAE and about 80\% source classification accuracy rate with notable source separation performance. 
 



\end{document}